\documentclass[journal]{IEEEtran}

\renewcommand\thesubsubsectiondis{\alph{subsubsection})}

\usepackage{titlesec}

\titleformat{\subsubsection}[block]
  {\itshape}                      
  {\thesubsubsectiondis}             
  {1em}                           
  {}                              
\titlespacing*{\subsubsection}
  {0pt}{0.5\baselineskip}{0.5\baselineskip}  

\usepackage[scr=boondox]{mathalfa}
\usepackage{pdflscape}
\usepackage[nocomma]{optidef} 
\usepackage{lipsum}
\usepackage[dvipsnames]{xcolor}
\usepackage[normalem]{ulem}
\usepackage[utf8]{inputenc}
\usepackage{multirow, multicol, makecell, booktabs, graphicx}
\usepackage{afterpage}
\usepackage{placeins}     
\usepackage{dsfont}
\usepackage{adjustbox}
\usepackage{url}
\usepackage[utf8]{inputenc}
\usepackage{amsmath,bm}
\usepackage{hyperref} 
\hypersetup{
     colorlinks   = true,
     citecolor    = blue,
     linkcolor = blue,
     urlcolor = blue
}
\usepackage[switch]{lineno}
\usepackage{bbm}
\usepackage{amsthm}
\usepackage{enumerate}
\usepackage{float}
\usepackage{stfloats}
\usepackage[numbers]{natbib}
\setcitestyle{numbers,square}
\usepackage[english]{babel}
\usepackage{cases}
\usepackage{array}
\usepackage{amsthm}
\usepackage{amsmath}
\usepackage{amssymb}

\theoremstyle{definition}

\theoremstyle{definition}

\theoremstyle{definition}

\theoremstyle{definition}

\theoremstyle{definition}

\usepackage{braket}
\usepackage{textgreek}
\usepackage{mathtools, cuted}
\usepackage{amsfonts}
\usepackage{algorithm}
\usepackage[noend]{algpseudocode}
\usepackage{hhline}
\usepackage{float}
\usepackage[normalem]{ulem} 
\usepackage{booktabs}
\usepackage{tabularx}

\algnewcommand\algorithmicinput{\textbf{Input:}}
\algnewcommand\Input{\item[\algorithmicinput]}
\algnewcommand\algorithmicoutput{\textbf{Output:}}
\algnewcommand\Output{\item[\algorithmicoutput]}

\usepackage [english]{babel}
\usepackage [autostyle, english = american]{csquotes}
\MakeOuterQuote{"}
\usepackage{hhline}
\usepackage{enumitem}
\usepackage{float}

\usepackage{caption}

\usepackage{subcaption}
\usepackage{amssymb}
\usepackage{siunitx}
\usepackage{tabstackengine}

\newcommand\AddLabel[1]{%
  \refstepcounter{equation}
  (\theequation)
  \label{#1}
}
\newcolumntype{C}{>{\hfil$\displaystyle}c<{$\hfil}}
\newcolumntype{R}{>{\hfil$\displaystyle}r<{$}}
\newcolumntype{L}{>{$\displaystyle}l<{$\hfil}}

\newcolumntype{Z}{>{\hfill$\displaystyle}X<{$\hfill}}
\newcolumntype{E}{>{\hfill$\displaystyle}X<{$}}
\newcolumntype{K}{>{$\displaystyle}X<{$\hfill}}

\usepackage{mathtools}

\newfloat{model}{tbp}{lom}
\floatname{model}{Model}

\newcounter{submodel}[model]
\newfloat{submodel}{tbp}{losm}
\floatname{submodel}{Model}

\newcommand{\integers}{\mathbb{Z}}

\newcommand{\Horizon}{\mathcal{H}}

\newcommand{\Jobs}{\mathcal{J}}

\newcommand{\Machines}{\mathcal{M}}

\newcommand{\setupMatrix}{\mathbb{S}}

\newcommand{\jm}{{jm}}

\newcommand{\fj}{{f_j}}

\newcommand{\ffp}{{ff'}}
\newcommand{\of}{{0f}}
\newcommand{\ofj}{{0,\fj}}
\newcommand{\og}{{0g}}
\newcommand{\fg}{{fg}}

\newcommand{\Families}{\mathcal{F}}

\newcommand{\machineState}{{\mathfrak{f}}}

\newcommand{\stime}{\tau}  
\newcommand{\ptime}{\delta}  
\newcommand{\weight}{\omega}  
\newcommand{\releaseTime}{r}  

\newcommand{\xh}{{\hat{x}}}

\newcommand{\pp}[1]{\left(#1\right)}

\newcommand{\EndOf}{\texttt{endOf}}
\newcommand{\PresenceOf}{\texttt{presenceOf}}

\newcommand{\EndBeforeEnd}{\texttt{endBeforeEnd}}

\newcommand{\StartBeforeStart}{\texttt{startBeforeStart}}
\newcommand{\NoOverlap}{\texttt{noOverlap}}

\newcommand{\Alternative}{\texttt{alternative}}
\newcommand{\StateFunction}{\texttt{state}}

\newcommand{\Permutation}{\text{Perm}}
\newcommand{\pulse}{\text{pulse}}
\newcommand{\AlwaysEqual}{\texttt{alwaysEqual}}
\newcommand{\AlwaysIn}{\texttt{alwaysIn}}

\usepackage{accents}

\begin{document}

\title{An Aligned Constraint Programming Model For Serial Batch Scheduling With Minimum Batch Size}

\author{
        Jorge A. Huertas,
        Pascal Van Hentenryck
    \thanks{
            This research was partly supported by the NSF \href{https://www.ai4opt.org}{\textit{AI Institute for Advances in Optimization}} (Award 2112533) \textit{(Corresponding author: Jorge A. Huertas.)}
    }%
    \thanks{
            J.A. Huertas, and P. Van Hentenryck are with the \href{https://www.gatech.edu}{\textit{Georgia Institute of Technology}}, Atlanta, GA, USA (e-mails: \href{mailto:huertas.ja@gatech.edu}{huertas.ja@gatech.edu}, \href{mailto:pvh@gatech.edu}{pvh@gatech.edu}).
    }
}



\maketitle

\begin{abstract}
In serial batch (s-batch) scheduling, jobs from similar families are grouped into batches and processed sequentially to avoid repetitive setups that are required when processing consecutive jobs of different families.  
Despite its large success in scheduling, only three Constraint Programming (CP) models have been proposed for this problem considering minimum batch sizes, which is a common requirement in many practical settings, including the ion implantation area in semiconductor manufacturing.
These existing CP models rely on a predefined virtual set of possible batches that suffers from the curse of dimensionality and adds complexity to the problem. 
This paper proposes a novel CP model that does not rely on this virtual set. Instead, it uses key alignment parameters that allow it to reason directly on the sequences of same-family jobs scheduled on the machines, resulting in a more compact formulation. This new model is further improved by exploiting the problem's structure with tailored search phases and strengthened inference levels of the constraint propagators.
The extensive computational experiments on nearly five thousand instances compare the proposed models against existing methods in the literature, including mixed-integer programming formulations, tabu search meta-heuristics, and CP approaches. The results demonstrate the superiority of the proposed models on small-to-medium instances with up to 100 jobs, and their ability to find solutions up to 25\% better than the ones produces by existing methods on large-scale instances with up to 500 jobs, 10 families, and 10 machines.
\end{abstract}



\section{Introduction} \label{sec: introduction}

In the current and highly competitive landscape of the manufacturing industry, companies are under growing pressure to minimize production costs and reduce cycle times. A crucial approach to achieving these goals and boosting production efficiency is processing multiple similar jobs in groups called \textit{batches} \cite{Monch2011-Survey}. Two types of batching can be distinguished in the scheduling literature, depending on how the jobs are processed inside their batch: (i) parallel batching (p-batch), where jobs inside a batch are processed in parallel at the same time \cite{Fowler2022-SurveyP-Batching}; and (ii) serial batching (s-batch), where jobs inside a batch are processed sequentially, one after the other \cite{PottsKovalyov2000-BatchSchedulingReview}. The benefits of p-batching in the manufacturing industry are straightforward due to the parallelized processing of the jobs inside a batch. In contrast, the benefits of s-batching usually come from grouping jobs that require similar machine configurations to avoid repetitive setups \cite{Wahl2024}. 

The s-batch problem falls under what is known in the scheduling literature as a \textit{family scheduling model} \cite{PottsKovalyov2000-BatchSchedulingReview}. In this framework, each job belongs to a specific family, which typically represents a shared machine setup or product recipe. Jobs are processed one after another on each machine, and switching between families often requires setup times, for example, cleaning the machine or reconfiguring it with the appropriate setup for the next job \cite{Wahl2024}. To reduce these costly transitions, batches are usually formed using jobs from the same family, thereby minimizing unnecessary setup operations \cite{Monch2011-Survey}.

S-batching is a common problem that appears in many manufacturing processes such as metal processing \cite{Gahm2022}, additive manufacturing (3D printing) \cite{Gahm2022, Uzunoglu2023}, paint manufacturing \cite{Shen2012}, pharmaceutical manufacturing \cite{Awad2022}, chemical manufacturing \cite{Karimi1995}, semiconductor manufacturing (SM) \cite{Monch2004, SMT2020-Paper}, and many more. Specifically, in the semiconductor manufacturing (SM) process, integrated circuits (ICs) are built on silicon wafers by repeating multiple layers of diffusion, photolithography, etching, ion implantation, and planarization operations \cite{Monch2013-FabBook, Chiang2012}. While p-batching is typically studied in the diffusion operations \cite{Ham2017-CP-p-batch-incompatible, Huertas2025_pbatching_TSM}, s-batching usually appears in the photolithography \cite{Monch2011-Survey} and ion implantation operations \cite{SMT2020-Paper}. 

Photolithography involves transferring the circuit designs from a photo-mask onto the wafers using ultraviolet light \cite{Akcah2001}. This process is repeated multiple times in different layers of the IC. The photolithography area is one of the main bottlenecks in a wafer fab due to the high cost and limited availability of exposure tools (i.e., steppers) \cite{Monch2011-Survey} and the rare gases involved in the process, such as neon, argon, and fluorine, used by excimer lasers \cite{Ebert2017NeonRecycle}. Furthermore, the recent conflict between Russia and Ukraine has reduced the availability of these rare gases \cite{Piotrowski2023CriticalRawMaterials}, since the neon produced by these two countries accounts for 70\% of the global market \cite{Chen2025}. In the photolithography area, the wafer lots (jobs) requiring the same photo-mask belong to the same family, and processing consecutive jobs from different families requires reticle changes \cite{Monch2004}. Hence, grouping jobs in batches to minimize these setups is beneficial for the process. Nonetheless, due to the limited availability of the rare gases involved, setting upper and lower bounds on the batch sizes helps prevent wasteful changes.

Ion implantation involves adding dopant ions into a silicon wafer by accelerating them through an electric field, ultimately altering the electrical properties of the wafers \cite{Lecuyer2009}. The scheduling problem in this area is usually tackled using dispatching rules that take into consideration the required setups \cite{SMT2020-Paper, Winkler2017, Chiang2012}. Nonetheless, it can also be addressed with the \textit{family scheduling model} by letting job families be determined by the distinct recipes of the jobs, which dictate unique configurations of ion source (e.g., germanium or phosphor gases), ion energy, dose, and acceleration voltages \cite{Winkler2017}. Jobs inside a batch are processed sequentially, and sequence-dependent setup times are required between consecutive batches of different families \cite{Duwayri2006}. These setups include changing the implant gas, which is costly and very time consuming \cite{SMT2020-Paper}. For this reason, the dispatching rules used in practice enforce a minimum batch size before conducting another setup, as evidenced in the instances of the Semiconductor Manufacturing Testbed (SMT) 2020 \cite{SMT2020-DataSpecification}.

In many practical applications, minimum batch sizes can be treated as soft constraints in the optimization model. In such settings, the cost of setting up a machine for a new job family is explicitly accounted for in the objective function, and the optimization implicitly determines the effective minimum batch size by balancing setup costs against other performance measures. In contrast, the \textit{Scheduling} stage in the semiconductor manufacturing planning matrix \cite{SurveySMSC2018-Part1} controls the progress of work within the production facility given higher-level decisions taken at the \textit{Production Planning} and \textit{Master Planning} stages, which directly interact with \textit{Material Requirements Planning} \cite{SurveySMSC2018-Part3}. As a result, the objectives at the Scheduling stage focus on job cycle times \cite{Ham2017-CP-p-batch-incompatible}, overall throughput and on-time targets \cite{Dauzere-Peres2024-FJSSSurvey}, or even machine utilization, without directly modeling the economic cost of starting a batch. Instead, such costs are translated into operational rules at higher-level stages by specifying minimum batch sizes that are passed down as hard constraints to the Scheduling stage. From the scheduler's perspective, these constraints ensure that only batches with a sufficient workload are initiated to justify machine usage \cite{Sung1997-bLB}, as observed, for example, in the ion implantation area in semiconductor manufacturing \cite{SMT2020-DataSpecification}. Other applications where hard constraints on minimum batch sizes are required include the metal cutting industry and pharmaceutical manufacturing \cite{Gahm2022, Wahl2023, Wahl2024}.

Despite this practical consideration, only a few studies in the s-batch literature have considered minimum batch sizes \cite{Wahl2024}. Moreover, despite the large success of Constraint Programming (CP) in scheduling \cite{Laborie2018}, a single article in the literature has proposed the only existing CP models for s-batch with minimum batch sizes, outperforming existing exact methods \cite{Huertas2025_sbatching_ORP}. Nonetheless, these existing CP models require the creation of a virtual set of possible batches that only adds complexity to the problem as the instance size grows. This paper fills this gap by proposing an improved CP model for s-batch scheduling with minimum batch size that does not rely on the virtual set of possible batches. Instead, it brings ideas from the \textit{Aligned} model for p-batching \cite{Ham2017-CP-p-batch-incompatible, Huertas2025_pbatching_TSM} into the s-batching realm, resulting in a more compact formulation.  This new model is further improved with tailored search phases and strengthened inference levels of the constraint propagators. The paper conducts extensive computational experiments that compare the proposed models against existing methods in the literature, demonstrating the superiority of the proposed models and their ability to find better solutions in large instances with up to 500 jobs, 10 families, and 10 machines.

The remainder of this paper is organized as follows. Section~\ref{sec: lit review} presents a literature review on s-batch methods suitable for semiconductor manufacturing. Section~\ref{sec: contributions} clearly outlines the contributions of this paper. Section~\ref{sec: problem description} formally describes the problem addressed. Section~\ref{sec: cp model} presents the proposed CP model. Section~\ref{sec: experiments} discusses the computational experiments conducted and their results. Finally, Section~\ref{sec: conclusion} presents the conclusions and outlines future lines of research.

\section{Literature review} \label{sec: lit review}

CP is a powerful method for solving combinatorial problems by defining constraints that a solution must satisfy. In CP, variables are assigned values from their domains, and the system works to ensure that all constraints are met. The search process explores the solution space in the form of a tree, where fixing the values of variables at each node branches out potential solutions. A key aspect of CP is the use of constraint propagation, which reduces the search space by pruning values from variable domains that cannot satisfy the constraints. Backtracking occurs when a branch leads to an invalid solution, allowing the search to revert to an earlier state and explore different variable assignments \cite{van2006handbook}.

Scheduling is one of the most successful application areas of CP. It leverages \textit{interval variables} to represent tasks or operations over time. These interval variables encapsulate three key components: the start time, the end time, and the presence status of the interval (indicating whether the task is executed or not). The start time and the end times define the size of the interval. The presence attribute allows CP to model both mandatory and optional activities within a schedule. In particular, it plays a crucial role in handling optional tasks and alternative resource allocations. \textit{Interval sequence} variables allow to model sequences of tasks. CP also uses \textit{cumulative functions} to model the evolution of resource usage over time, ensuring that resource capacities are respected by aggregating the resource demands of all overlapping tasks. These cumulative functions are crucial in preventing over-utilization of resources like machines, personnel, or energy. The efficiency of the search process is further enhanced by \textit{global constraints} that exploit the structure of the problem to propagate information and prune variable domains efficiently. Together, CP and its tools for constraint propagation, domain pruning, backtracking, and cumulative functions provide a structured and efficient approach for solving complex scheduling problems \cite{van2006handbook, Laborie2018}. The interested reader on an in-depth CP overview is referred to the well-known book by \citet{van2006handbook}, and the paper by \citet{Laborie2018}.

The scheduling literature distinguishes two types of batching: p-batching and s-batching \cite{Fowler2022-SurveyP-Batching}. Existing approaches for p-batching in semiconductor manufacturing that consider multiple incompatible job families, multiple parallel machines, and non-identical release times include mixed-integer programming (MIP) models \cite{Cakici2013, Ham2017-CP-p-batch-incompatible}, Variable Neighborhood Search (VNS) heuristics \cite{Cakici2013}, and CP models \cite{Ham2017-CP-p-batch-incompatible, Huertas2025_pbatching_TSM}. The \textit{Aligned} model for p-batching (p-A) is a CP model that relies on key optional parameters inside a global constraint to align the start and end times of overlapping jobs of the same family, therefore creating parallel batches. This p-A model consistently outperforms the existing MIP models and VNS approaches \cite{Ham2017-CP-p-batch-incompatible, Huertas2025_pbatching_TSM}.

\begin{figure}[t]
    \centering
    \begin{subfigure}[b]{0.35\linewidth}
        \centering
        \includegraphics[width=\textwidth]{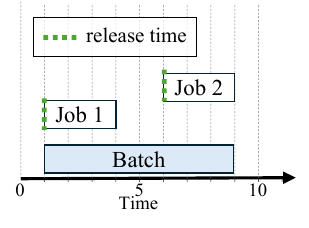}
        \caption{Preemptive}
        \label{fig: preemptive}
    \end{subfigure}
    \quad\quad\quad
    \begin{subfigure}[b]{0.35\linewidth}
        \centering
        \includegraphics[width=\textwidth]{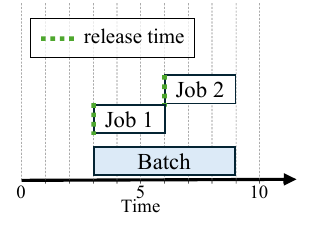}
        \caption{Non-preemptive}
        \label{fig: non-preemptive}
    \end{subfigure}
    \caption{Batch processing type}
    \label{fig: s-batch processing types}
\end{figure}

\begin{figure}[t]
    \centering
    \begin{subfigure}[b]{0.35\linewidth}
        \centering
        \includegraphics[width=\textwidth]{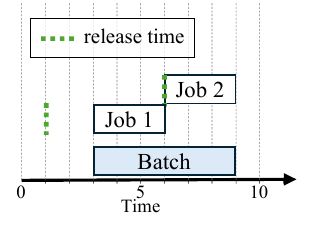}
        \caption{Flexible}
        \label{fig: flexible}
    \end{subfigure}
    \quad\quad\quad
    \begin{subfigure}[b]{0.35\linewidth} 
        \centering
        \includegraphics[width=\textwidth]{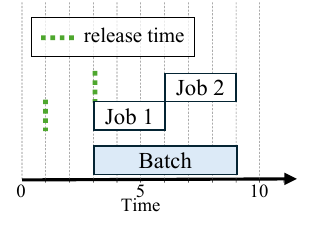}
        \caption{Complete}
        \label{fig: complete}
    \end{subfigure}
    \caption{Batch initiation type}
    \label{fig: s-batch initiation types}
\end{figure}

At the other end of the spectrum, the scheduling literature distinguishes multiple s-batch variations \cite{Wahl2024}. Two variations exist depending on the time when jobs are considered completed: under \textit{item availability}, the jobs become completed as soon as their processing time is finished; instead, under \textit{batch availability}, jobs are considered completed when the entire batch has been processed \cite{PottsKovalyov2000-BatchSchedulingReview}. Figure \ref{fig: s-batch processing types} shows two types of variations depending on whether idle times are allowed to preempt the processing of jobs inside a batch: \textit{Preemptive} processing allows idle times inside a batch, and \textit{non-preemptive} forbids them \cite{Jordan1996}. Figure \ref{fig: s-batch initiation types} shows the two types of s-batch variations depending on the batch initiation: \textit{Flexible initiation} allows the batches to start before the release time of one (or more) of its jobs, while a \textit{complete initiation} forces all the jobs in the batch to be released before the batch start time. This paper focuses on the s-batch variation that appears in the photolithography and ion implantation areas of a wafer fab, which considers item availability, preemptive processing, and flexible initiation \cite{Monch2011-Survey}. 
For this reason, this variation is henceforward referred to as the IPF variation.

Existing s-batch reviews in the literature include the early one by \citet{PottsKovalyov2000-BatchSchedulingReview} in 2000, the one by \citet{Monch2011-Survey} in 2011, and the most recent by \citet{Wahl2024} in 2024. In the most recent one, only seven articles are mentioned to consider minimum batch sizes \cite{Sung1997-bLB, Mosheiov2008-bLB, Chretienne2011-bLB, Hazir2014-bLB, Castillo2015-bLB, Shahvari2017-bLB, Shahvari2016-bLB, Huertas2025_sbatching_ORP}, and only one article after such review has addressed the problem with such minimum requirement \cite{Huertas2025_sbatching_ORP}.

Some early approaches to solve the problem include dynamic programming (DP) \cite{Sung1997-bLB, Chretienne2011-bLB, Hazir2014-bLB}, heuristic algorithms \cite{Sung1997-bLB}, and rounding approximations \cite{Mosheiov2008-bLB}. However, all these early attempts only focus on a single machine and assume identical release times of the jobs. The natural extension to consider multiple machines was addressed using Genetic Algorithms (GA), but relying on identical job release times \cite{Castillo2015-bLB}. The extension with multiple machines and non-identical job release times has been addressed using mixed-integer programming (MIP) models \cite{Shahvari2016-bLB, Shahvari2017-bLB}, Tabu Search (TS) \cite{Shahvari2016-bLB, Shahvari2017-bLB}, and, more recently, CP \cite{Huertas2025_sbatching_ORP}.

\citet{Shahvari2017-bLB} proposed an MIP model for the IPF variation that uses continuous variables for batch and job completion times and binary variables to define the relative position of every pair of batches and jobs. Because of this structure, their model is referred to as the \textit{Relative Positioning} (RP) MIP model \cite{Huertas2025_sbatching_ORP}. They report results on only 22 instances and, on their largest instance, a moderately loaded scenario with 27 jobs, 9 families, and 4 machines, the RP model hits a time limit of two days without proving optimality. To remedy this, they also propose a TS metaheuristic that can obtain the optimal solution within 20 minutes on their easiest instances and, within 3 hours, solutions up to 7\% better than those of the RP model on the largest instance. The same authors later embed this TS in a larger flowshop environment \cite{Shahvari2016-bLB}. Overall, both the RP model and the TS metaheuristic are evaluated only on relatively small instances, with at most 27 jobs and multi-hour runtimes, which is impractical for real wafer fabs that handle hundreds of jobs and where scheduling tools typically generate schedules every 5 to 10 minutes \cite{Ham2017-CP-p-batch-incompatible}.

Recently, \citet{Huertas2025_sbatching_ORP} proposed the first three CP models for s-batching with minimum batch sizes: the \textit{Interval Assignment} (IA), \textit{Global} (G), and \textit{Hybrid} (H) models. All three define interval variables for jobs and batches on the machines and assign job intervals to batch intervals; they differ mainly in how they enforce minimum batch sizes. The IA model uses linear constraints on the sum of job presence literals in each batch. Although these constraints correctly enforce minimum batch sizes, they remain local and provide limited guidance to the search. The G model instead uses global constraints over cumulative functions that capture the batch sizes, providing a more global view that strengthens propagation and prunes domains more effectively, at the cost of unnecessarily tracking batch size over time. The H model strikes a balance by combining the extra global constraints of the G model with the sum-of-presence constraints of the IA model. Their extensive computational experiments on over a thousand instances with up to instances with up to 100 jobs, 7 families, and 5 machines show that these CP models consistently outperform existing approaches, finding in just 10 minutes solutions that are up to 7\% better than those obtained after one hour with the RP model. Nonetheless, all three CP models rely on a virtual set of possible batches that suffers from the curse of dimensionality and increases model complexity as instance size grows.

\section{Contributions} \label{sec: contributions}

This paper proposes a new CP model for s-batch scheduling with minimum batch sizes that does not rely on the virtual set of possible batches required by the IA, G, and H models. It brings ideas from the p-A model for parallel batching into the serial batching realm. The resulting \textit{Aligned} model for s-batching (s-A) has a more compact formulation than the IA, G, and H models. We further enhance the s-A model by analyzing the structure of the formulation and guiding the CP engine toward appropriate search phases, which leads to the improved s-A* model. Extensive computational experiments on over one thousand small-scale instances with up to 100 jobs, 7 families, and 5 machines show that the proposed s-A and s-A* models consistently outperform the RP MIP model and the IA, G, and H CP models. Additional experiments on over three thousand large-scale instances with up to 500 jobs, 10 families, and 10 machines demonstrate that the proposed models with enhanced inference levels can produce solutions up to 25\% better than existing CP models.

\section{Problem description} \label{sec: problem description}

Let $\Jobs$ be the set of jobs and $\Machines$ be the set of machines. Jobs are partitioned into families $\Families$ based on their similarity. Let $f_j \in \Families$ be the family of job $j \in \Jobs$, and $\Jobs_f = \set{j \in \Jobs : f_j = f}$ be the subset of jobs that belong to family $f \in \Families$. Each job $j$ has a weight $\weight_j$, a release time $\releaseTime_j$, and a processing time $\ptime_j$. All the jobs can be scheduled on all the machines, and each machine can process only one job at a time. Consecutive jobs of the same family $f$ are a \textit{serial batch} and it is necessary to have a minimum and maximum number $l_f$ and $u_f$ of consecutive jobs in the batch before processing another batch ($0 < l_f \leq u_f$).  No setup is required between consecutive jobs of the same batch. However, let $\stime_\fg$ be the \textit{family setup time} when a batch of family $g \in \Families$ is immediately preceded by a batch of a different family $f \in \Families$, or $\stime_\og$ if there is no preceding batch. It is assumed that these setup times satisfy the triangular inequality, meaning that $\stime_\ffp \leq \stime_\fg + \stime_{gf'}$. 

\citet{Huertas2025_sbatching_ORP} demonstrated that it was significantly more difficult to optimize for the total weighted completion time (TWCT) than the makespan. For this reason, this paper focuses on minimizing the total weighted completion time (TWCT) of the jobs, which is the weighted sum of the job completion times. The constraints include selecting the machine where each job is processed and ensuring that jobs processed on the same machine do not overlap while respecting: the job release times, the family setup times between batches, and the minimum (and maximum) batch size requirements.

\section{Aligned CP model} \label{sec: cp model}

The IA, G, and H models rely on a virtual set of possible batches to assign job intervals to batch intervals. This allows them to count the number of jobs inside the activated batches and ultimately enforce the minimum batch sizes. However, defining this virtual set is already a complex task. If too few batches are defined, the problem may become infeasible; if too many are defined, the problem size increases dramatically. For this reason, the s-A model proposed in this paper does not use a virtual set of possible batches and instead reasons directly on consecutive jobs of the same family. We refer to any such sequence of consecutive same-family jobs on a machine as a \emph{family block}. Conceptually, a family block plays the same role as a serial batch. However, we purposely adopt this term to emphasize how the s-A model operates: family blocks are induced implicitly by the sequence of jobs, whereas batches in the IA, G, and H models rely on an explicit set of possible batch indices.

This section presents the mathematical formulation of the s-A model. The syntax follows that of IBM ILOG CPLEX CP Optimizer \cite{Laborie2018}, a widely used CP solver. The model is organized into three sections:
(i) the \textit{Core} section, which sequences jobs on machines;
(ii) the \textit{Family Block} section, which constructs family blocks from consecutive same-family jobs without explicit batch indices; and
(iii) the \textit{Sizing} section, which enforces minimum (and maximum) size requirements on the family blocks.

The Core section is a straightforward CP formulation, commonly presented in CP tutorials and official documentation \cite{IBM_CP_Optimizer_Sequence_Dependent_Setup_Times} to demonstrate how sequence-dependent setup times can be considered when sequencing non-overlapping jobs on machines. Setup times are stored in the matrix $\setupMatrix = \{\stime_{fg}\}_{f,g\in \Families} \in \integers^{\Families \times \Families}$, which is assumed to satisfy the triangle inequality. This requirement avoids inconsistencies in setup durations, such as indirect transitions being shorter than direct ones, which would undermine the logic of the sequencing decisions. The Core section schedules similar jobs consecutively to avoid unnecessary setups. These consecutive jobs can be viewed as a serial batch, so the Core section is itself a model for s-batching. Nonetheless, it cannot guarantee the minimum batch sizes.

To address this limitation, the s-A model extends the Core section with the Family Block and Sizing sections. The Family Block section introduces additional virtual interval variables for the jobs that extend over the full duration of their family block. This causes virtual job intervals in the same family block to overlap, which in turn allows the Sizing section to define cumulative functions over these overlapping intervals to capture the size of the blocks and enforce their minimum and maximum size requirements.

To illustrate how each section of the model contributes to constructing the solution, consider the simple example in Table~\ref{tab: example}, which involves 5 jobs belonging to 2 families and a single machine. Assume the minimum batch sizes are $l_1 = 3$ for family 1 and $l_2 = 2$ for family 2. Also, let the initial setup times be $\tau_{0,1} = \tau_{0,2} = 1$, and the setup times between families be $\tau_{1,2} = \tau_{2,1} = 3$. The following sections describe each part of the model in detail and illustrate how the solution of this example evolves as each section is incrementally added.

\begin{table}[t] 
    \centering
    \caption{Data for the illustrative example}\label{tab: example}
    \begin{tabular}{cccccc}
        \toprule
        \textbf{Job} & \textbf{Weight} & \textbf{Release Time} & \textbf{Processing Time} & \textbf{Family} \\
        \midrule
        1 & 1 & 1  & 2 & 1 \\
        2 & 1 & 5  & 2 & 1 \\
        3 & 1 & 6  & 2 & 2 \\
        4 & 1 & 12 & 2 & 2 \\
        5 & 1 & 11 & 2 & 1 \\
        \bottomrule
    \end{tabular}
\end{table}

\subsection{Core Section}

\setcounter{model}{1}
\begin{subequations}\label{eq:CP}
\begin{submodel}[thbp]
\caption{Core section} \label{model:core}
\renewcommand{\arraystretch}{1.2}
\begin{tabularx}{\linewidth}{@{}cL@{}L@{}ER@{}}
    \multicolumn{5}{@{}l}{\textbf{Variables and functions}:} \\
    \multicolumn{5}{@{}l}{\textbullet ~ Interval variables:} \\
    & x_j & \multicolumn{3}{@{}K}{ \in \set{[s,s + \ptime_j): s \in \Horizon, s \geq \max \{ \releaseTime_j, \stime_\ofj \} },} \\
    & & \multicolumn{2}{@{}R}{\forall ~j \in \Jobs;} & \AddLabel{eq: cp def - interval job} \\
    & x_\jm & \multicolumn{3}{@{}K}{ \in \set{[s,s + \ptime_j) : s \in \Horizon } \cup \set{\perp}} \\
    &  & \multicolumn{2}{@{}R}{\forall ~ j \in \Jobs, m \in \Machines;} & \AddLabel{eq: cp def - interval job on machine}\\
    \multicolumn{5}{@{}l}{\textbullet ~ Sequence variables:} \\
    & \varphi_m & \multicolumn{3}{@{}K}{\in \Permutation(\set{x_\jm}_{j \in \Jobs}) \text{ with types }\set{f_j}_{j \in \Jobs},} \\
    & & \multicolumn{2}{@{}R}{\forall ~m \in \Machines;} & \AddLabel{eq: cp def - job sequence}\\
    \multicolumn{5}{@{}l}{\textbullet ~ State functions:} \\
    & \machineState_m &: \StateFunction & \forall ~ m \in \Machines; & \AddLabel{eq: cp def - state} \\
\end{tabularx}
\begin{tabularx}{\linewidth}{@{}cR@{}LER@{}}
    \hline
    \multicolumn{5}{@{}l}{\textbf{Formulation}:} \\
    & \multicolumn{3}{l}{\(\text{minimize~} \displaystyle\sum_{j \in \Jobs} \weight_j \cdot \EndOf(x_j)\)} & \AddLabel{eq: cp - obj function item} \\
    \multicolumn{5}{l}{subject to,}\\
    & \multicolumn{2}{L}{\Alternative(x_j, \set{x_\jm}_{m \in \Machines}),} & \forall ~ j \in \Jobs; & \AddLabel{eq: cp - job in one machine}\\
    & \multicolumn{2}{L}{\NoOverlap(\varphi_m, \setupMatrix), }& \forall ~ m \in \Machines; & \AddLabel{eq: cp - no overlap jobs}\\
    & \multicolumn{3}{L}{\AlwaysEqual(\machineState_m, x_\jm, f_j), \quad \forall ~ j \in \Jobs, m \in \Machines;} & \AddLabel{eq: cp - state is job family} \\
\end{tabularx}
\end{submodel}

\begin{submodel}[thbp!]
\caption{Family block section} \label{model:family_block}
\renewcommand{\arraystretch}{1.2}
\begin{tabularx}{\linewidth}{@{}cL@{}L@{}ER@{}}
    \multicolumn{5}{@{}l}{\textbf{Additional variables}:} \\
    \multicolumn{5}{@{}l}{\textbullet ~ Interval variables:} \\
    & \xh_\jm & \multicolumn{3}{@{}L}{ \in \set{[s, e) : s,e \in \Horizon, s \leq e} \cup \set{\perp},}\\
    & & \multicolumn{2}{@{}E}{ \forall ~ j \in \Jobs, m \in \Machines;} & \AddLabel{eq: cp def - virtual job on machine}\\
\end{tabularx}
\begin{tabularx}{\linewidth}{@{}cR@{}LER@{}}
    \hline
    \multicolumn{5}{@{}l}{\textbf{Additional constraints}:} \\
    & \multicolumn{4}{L}{\PresenceOf(x_\jm) = \PresenceOf (\xh_\jm),} \\
    & & \multicolumn{2}{@{}E}{\forall ~ j \in \Jobs, m \in \Machines;} & \AddLabel{eq: cp - presence equality} \\
    & \multicolumn{4}{L}{\StartBeforeStart(\xh_\jm, x_\jm),} \\
    & & \multicolumn{2}{@{}E}{\forall ~ j \in \Jobs, m \in \Machines;} & \AddLabel{eq: cp - start before start} \\
    & \multicolumn{4}{L}{\EndBeforeEnd(x_\jm, \xh_\jm),} \\
    & & \multicolumn{2}{@{}E}{\forall ~ j \in \Jobs, m \in \Machines;} & \AddLabel{eq: cp - end before end} \\
    & \multicolumn{4}{L}{\AlwaysEqual(\machineState_m, \xh_\jm, f_j, \texttt{True}, \texttt{True}),} \\
    & & \multicolumn{2}{@{}E}{\forall ~ j \in \Jobs, m \in \Machines;} & \AddLabel{eq: cp - alignment} \\
\end{tabularx}
\end{submodel}

\begin{submodel}[thbp!]
\caption{Sizing section} \label{model:batchsizeglobal}
\renewcommand{\arraystretch}{1.2}
\begin{tabularx}{\linewidth}{@{}cL@{}L@{}ER@{}}
    \multicolumn{5}{@{}l}{\textbf{Additional functions}:} \\
    \multicolumn{5}{@{}l}{\textbullet ~ Cumulative functions:} \\
    & n_{mf} &= \sum_{j \in \Jobs_f} \pulse \pp{\xh_\jm, 1}, & \forall ~ m \in \Machines, f \in \Families. & \AddLabel{eq: cd  def - cumul}\\
\end{tabularx}
\begin{tabularx}{\linewidth}{@{}cR@{}LER@{}}
    \hline
    \multicolumn{5}{@{}l}{\textbf{Additional constraints}:} \\
    & \multicolumn{4}{L}{\AlwaysIn(n_{mf}, \xh_\jm, l_f, u_f),} \\
    & & \multicolumn{2}{@{}E}{\forall ~ m \in \Machines, f \in \Families, j \in \Jobs_f.} & \AddLabel{eq: cp - always in}\\
\end{tabularx}
\end{submodel}
\end{subequations}

Let $\Horizon = \set{0, 1, \ldots, H}$ be the scheduling horizon, where $H = \max\{ \max_{j \in \Jobs} \releaseTime_j, \max_{f \in \Families} \stime_\of \} + (\sum_{j \in \Jobs} \ptime_j) + (|\Families| - 1) \cdot \max_{f,g \in \Families} \stime_\fg$ is the horizon's upper bound. This upper bound assumes that all jobs are processed sequentially on a single machine after all of them have already been released and paid the initial setup time. Thus, all the jobs of the same family are scheduled sequentially and at most $|\Families| -1$ setups between families need to be made.

Model \ref{model:core} presents the Core section of the s-A model. Equation \eqref{eq: cp def - interval job} defines an interval variable of the form $x_j = [s,s + \ptime_j)$, i.e., it has a size of exactly $\ptime_j$ units of time, which represents job $j \in \Jobs$. This interval can only start after the job's $\releaseTime_j$ and its family's initial setup time $\stime_\ofj$. Equation \eqref{eq: cp def - interval job on machine} defines an optional interval variable $x_\jm$ of size $\ptime_j$ that represents the option of job $j$ being processed on machine $m \in \Machines$. This interval is optional since it is allowed to take the value $\perp$, which indicates its absence from the solution. It is not necessary to indicate that variable $x_\jm$ can only start after $\releaseTime_j$ because, if selected to be present, this variable is going to be synchronized with variable $x_j$, which already accounts for this. Equation \eqref{eq: cp def - job sequence} defines a sequence variable $\varphi_m$ of jobs on machine $m$, which is a permutation of the job intervals $\set{x_\jm}_{j \in \Jobs}$ on such machine, whose types are the associated job families. These types serve as row and column identifiers in the setup times matrix $\setupMatrix$ to ensure the minimum time distance between consecutive job intervals in the sequence. The last element of the core model is defined by equation \eqref{eq: cp def - state}, which defines a state variable $\machineState_m$ that represents the family being processed on machine $m$. 

Objective function \eqref{eq: cp - obj function item} minimizes the TWCT under item availability, tallying the completion of the jobs as soon as their processing time finishes.  Constraints \eqref{eq: cp - job in one machine} use the $\Alternative(v, V)$ global constraint, which receives an interval variable $v$ and a set of optional intervals $V$. This global constraint ensures that if the interval $v$ is present in the solution, then exactly one interval from the set $V$ is selected to be present in the solution as well, and synchronizes it with interval $v$. Hence, constraints \eqref{eq: cp - job in one machine} ensure that each job is processed on exactly one machine. Constraints \eqref{eq: cp - no overlap jobs} use the $\NoOverlap(\varphi, \setupMatrix)$ global constraint, which receives an interval sequence variable $\varphi $ and a matrix with transition times $\setupMatrix$. This global constraint ensures non-overlapping intervals in the sequence defined by the permutation $\varphi$, with a minimum distance between them given by the transition times $\setupMatrix$. Hence, constraints \eqref{eq: cp - no overlap jobs} ensure that only one job is processed at a time on each machine, while respecting the family setup times. Constraints \eqref{eq: cp - state is job family} use the $\AlwaysEqual(h, v, a)$ global constraint, which receives a state function $h$, an interval variable $v$, and an integer value $a$. This global constraint ensures that if $v$ is present in the solution, then the state function takes a constant value $h(t) = a$ at any point in time during interval $v$, i.e., $t\in v$. Hence, constraints \eqref{eq: cp - state is job family} ensure that the sate of each machine is the family of the job being processed.

\begin{figure}[t!]
    \centering
    \includegraphics[width=\linewidth]{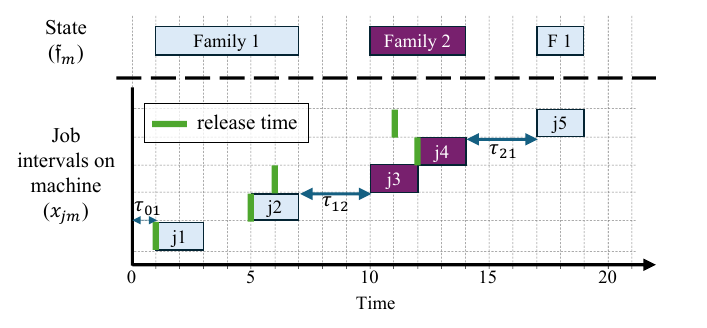}
    \caption{Solution of the Core section}
    \label{fig:core solution}
\end{figure}

The solution produced by the Core section for the example in Table~\ref{tab: example} is illustrated in Figure~\ref{fig:core solution}. Since there is only one machine, the Gantt chart represents its schedule. The figure shows how the job-on-machine intervals $x_\jm$ are assigned their optimal values, achieving a total weighted completion time (TWCT) of 55. This value corresponds to the sum of the completion times of all jobs, given that each job has a weight of 1.
Each job is shown as a box, with its color indicating its family: blue for family 1 and purple for family 2. Each job appears on a separate vertical level, and the green vertical lines mark their respective release times. The initial setup time is applied at time 0, allowing job 1 to start as soon as it is released. From there, jobs are scheduled sequentially without overlapping. Setup times between families are respected—for example, between jobs 2 and 3, and again between jobs 4 and 5.
Although the minimum batch size for family 1 is $l_1 = 3$, the Core solution only places two consecutive jobs of that family (jobs 1 and 2) before switching to jobs of family 2. Job 5 from family 1 is scheduled last to minimize TWCT. Thus, the Core section effectively schedules non-overlapping jobs while respecting setup times between families. However, it cannot enforce minimum batch size requirements. For this reason, the Family Block and Sizing sections are needed.

\subsection{Family block section}

The Family Block section in Model~\ref{model:family_block} creates virtual intervals for the jobs that extend through the whole duration of their corresponding family block. Hence, for jobs belonging to the same family block, their virtual intervals not only overlap but are also aligned. Equation~\eqref{eq: cp def - virtual job on machine} defines an optional virtual interval variable for job $j$ on machine $m$, which is intended to capture the entire duration of the family block in which job $j$ is processed. While $x_\jm$ represents the actual processing interval of job $j$ on machine $m$, the hat symbol $\xh_\jm$ denotes its virtual interval. The start and end times of these virtual intervals are not fixed, allowing them to contain their corresponding job interval and enabling other jobs in the same family block to precede or follow job $j$.

Constraints~\eqref{eq: cp - presence equality} link the virtual and actual intervals by enforcing that, whenever a job is assigned to a machine, both $x_\jm$ and $\xh_\jm$ are present on that machine. Since the virtual intervals represent the whole duration of the corresponding family block, the job intervals must be contained within them. Constraints~\eqref{eq: cp - start before start} use the global constraint $\StartBeforeStart(v_1, v_2)$, which receives two optional interval variables $v_1$ and $v_2$ and, if both are present, ensures that $v_1$ starts no later than $v_2$. Similarly, constraints~\eqref{eq: cp - end before end} use the global constraint $\EndBeforeEnd(v_1, v_2)$, which, when both intervals are present, ensures that $v_1$ ends no later than $v_2$. Together, constraints~\eqref{eq: cp - start before start} and~\eqref{eq: cp - end before end} ensure that each job interval lies within its corresponding virtual interval.

Finally, Constraints~\eqref{eq: cp - alignment} use the additional optional alignment parameters $align\_s$ and $align\_e$ inside the $\AlwaysEqual(h, v, a, align\_s, align\_e)$ global constraint. By default, these parameters are disabled; when enabled, they align the start and end times of the interval $v$ with the start and end times of the time segment where the state function $h$ takes the value $a$. In this way, Constraints~\eqref{eq: cp - alignment} ensure that the virtual intervals of all the jobs in the same family block are aligned, so that they can later be aggregated in the Sizing section.

\begin{figure}[t!]
    \centering
    \includegraphics[width=\linewidth]{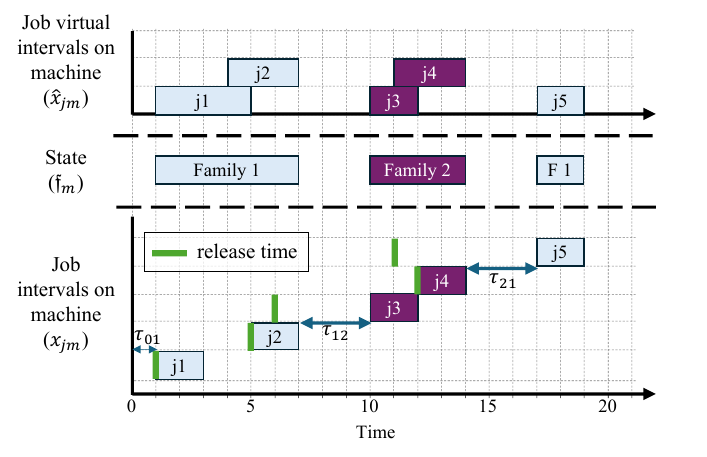}
    \caption{Possible solution of the joint Core and Family Block sections with alignment parameters disabled}
    \label{fig:family block solution alignment off}
\end{figure}

\begin{figure}[t!]
    \centering
    \includegraphics[width=\linewidth]{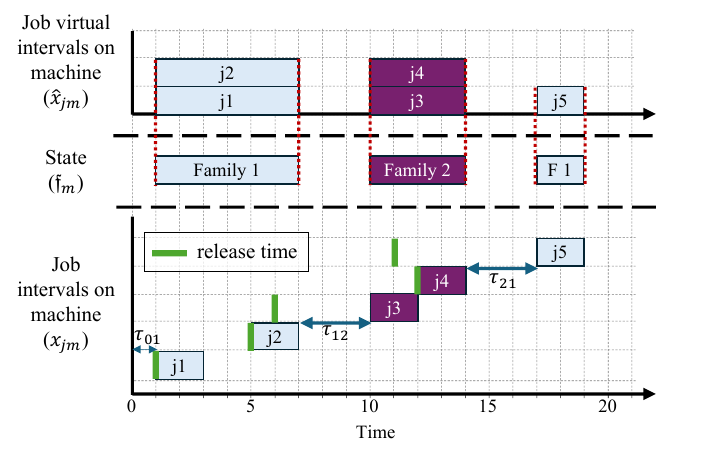}
    \caption{Solution of the joint Core and Family Block sections with alignment parameters enabled}
    \label{fig:family block solution alignment on}
\end{figure}

Coming back to the illustrative example, Figures~\ref{fig:family block solution alignment off} and~\ref{fig:family block solution alignment on} show two possible solutions of the joint Core and Family Block sections. Figure~\ref{fig:family block solution alignment off} displays a possible solution with the alignment parameters disabled, whereas Figure~\ref{fig:family block solution alignment on} shows the solution with them enabled. Note that the solution itself does not change with respect to the one of the independent Core section shown in Figure~\ref{fig:core solution}; instead, additional virtual intervals are introduced in the top Gantt chart. When the alignment parameters are disabled, these virtual intervals do not fully capture the duration of the corresponding family blocks. In contrast, when the alignment parameters are enabled, the virtual intervals of all the jobs within the same family block are aligned, as highlighted by the vertical red dotted lines in Figure~\ref{fig:family block solution alignment on}. Nonetheless, the Family Block section alone does not yet enforce minimum batch-size requirements; this is handled by the Sizing section.

\subsection{Sizing section}

The Sizing section in Model~\ref{model:batchsizeglobal} is responsible for enforcing the minimum (and maximum) size requirements on the family blocks. To achieve this, equation~\eqref{eq: cd  def - cumul} introduces a cumulative function $n_{mf}$ which tallies the number of overlapping virtual intervals of jobs from family $f \in \Families$ on machine $m \in \Machines$. To do so, it pulses one unit throughout the whole duration of the virtual interval of each job of family $f$. Since Model~\ref{model:family_block} aligns the virtual intervals of jobs in the same family block, the cumulative function effectively captures the size of the active family block at any point in time.

Constraints~\eqref{eq: cp - always in} use the $\AlwaysIn(n,v,l,u)$ global constraint, which receives a cumulative function $n$, an interval variable $v$, and integer bounds $l \leq u$. This global constraint ensures that, if $v$ is present in the solution, then the value of $n$ remains within the interval $[l,u]$ at any time during $v$, i.e., $l \leq n(t) \leq u$ for all $t \in v$. Hence, Constraints~\eqref{eq: cp - always in} guarantee that, whenever a virtual interval $\xh_\jm$ is present, the corresponding family block has a size between $l_f$ and $u_f$. In particular, the lower bound $l_f$ enforces the required minimum batch size for family $f$, ensuring that any family block that appears in the schedule is large enough to satisfy the minimum batch-size requirement.

\begin{figure}[t!]
    \centering
    \includegraphics[width=\linewidth]{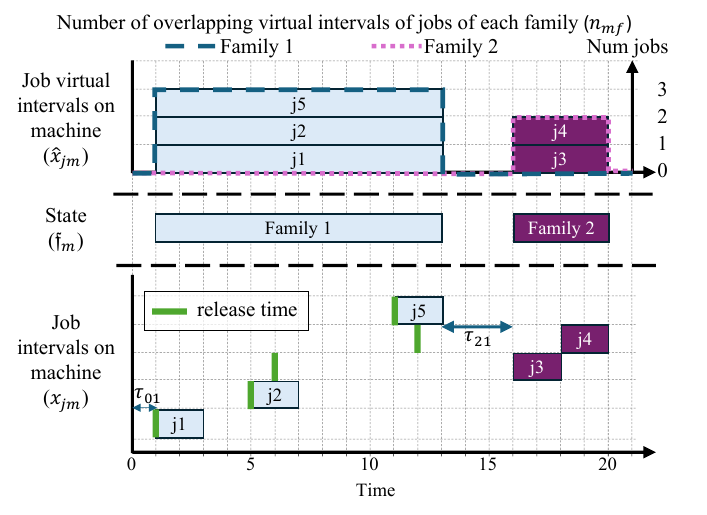}
    \caption{Solution of the s-A model (joint Core, Family Block, and Sizing sections)}
    \label{fig:sizing solution}
\end{figure}

Coming back to the illustrative example, the inclusion of the Sizing section in the model leads to a noticeable change in the solution, as the minimum batch size requirements are now enforced. Figure~\ref{fig:sizing solution} presents the complete solution obtained by the combined Core, Family Block, and Sizing sections. To satisfy the minimum batch size $l_1 = 3$ for family~1, the model must delay the processing of jobs~3 and~4 (from family~2) and schedule job~5 earlier. As a result, the first family block of family~1 now includes jobs~1, 2, and~5, which meets the minimum batch size requirement $l_1 = 3$. Similarly, the family block of family~2 includes jobs~3 and~4, satisfying the minimum batch size requirement $l_2 = 2$. These constraints are enforced by the Sizing section through the cumulative functions $n_{mf}$ and the $\AlwaysIn$ global constraints in \eqref{eq: cd  def - cumul}–\eqref{eq: cp - always in}, which rule out any family block whose size is below the required threshold. Enforcing these minimum batch sizes impacts the objective function, increasing the TWCT from 55 to 61.

\subsection{Improved model} \label{sec: improved model}

The s-A model can be further enhanced by subtle but powerful changes to the formulation and by exploiting its structure in the search process, resulting in the improved s-A* model. The key differences between the s-A and s-A* models are presented below.

\subsubsection{Modifications to the formulation}

Constraints~\eqref{eq: cp - always in} are critical in the s-A formulation because they ensure that the minimum batch sizes are respected. They guarantee that, whenever the virtual interval $\xh_\jm$ is present, the size of the corresponding family block of family $f$ on machine $m$ satisfies $l_f \leq n_{mf}(t) \leq u_f$ at every point in time $t \in \xh_\jm$. Because of constraints~\eqref{eq: cp - presence equality}–\eqref{eq: cp - end before end}, the virtual interval $\xh_\jm$ is always present whenever the interval $x_\jm$ is present, which is contained within $\xh_\jm$; moreover, all virtual intervals $\xh_\jm$ in the same family block are aligned by constraints~\eqref{eq: cp - alignment}. As a consequence, the same bounds on $n_{mf}(t)$ can be enforced at every point in time during interval $x_\jm$ without changing the feasible set. Therefore, constraints~\eqref{eq: cp - always in} can be replaced by constraints~\eqref{eq: cp def - always in 2}, which simply use $x_\jm$ instead of $\xh_\jm$ in the $\AlwaysIn$ global constraint. This modification moves the propagation of the family-block size bounds from the virtual intervals to the actual job intervals, so that the batch-size requirements can start pruning the search tree as soon as the core variables $x_\jm$ are decided, without waiting for the virtual intervals to be fixed.
\begin{align}
    \AlwaysIn &(n_{mf}, x_\jm, l_f, u_f), & & \forall ~ m \in \Machines, f \in \Families, j \in \Jobs_f. \label{eq: cp def - always in 2}
\end{align}
\subsubsection{Search phases}

The s-A* model also exploits the structure of the formulation through dedicated search phases. The idea is to first assign the core decision variables that determine the sequencing and timing of jobs on the machines, and only then assign the virtual intervals used to construct family blocks and enforce sizing.

The two search phases used in s-A* are:
\begin{itemize}
    \item Phase 1: $\{x_\jm\}_{j \in \Jobs,\, m \in \Machines} \cup \{x_j\}_{j \in \Jobs}$.
    \item Phase 2: $\{\xh_\jm\}_{j \in \Jobs,\, m \in \Machines}$.
\end{itemize}
In Phase~1, the solver focuses on the core intervals $x_\jm$ and $x_j$, which define the assignment of jobs to machines and their start and end times, and thus determine the underlying schedule. In Phase~2, once this structure is fixed, the solver assigns the virtual intervals $\xh_\jm$, which inherit the family-block structure from the Core and Family Block sections and are then used in the Sizing section to enforce the minimum batch-size requirements. This ordering guides the search to first resolve the main combinatorial decisions and then refine the auxiliary structure, leading to substantial computational gains in practice.

\section{Computational experiments} \label{sec: experiments}

This section divides the computational experiments in two parts. \S~\ref{sec: initial comparison} compares the proposed s-A and s-A* models against the existing MIP and CP models using the exact same small-to-medium instances from \citet{Huertas2025_sbatching_ORP}. Then, \S~\ref{sec: large-scale results} presents new large-scale instances to compare the proposed models (under different inference levels of the constraint propagators) against the existing CP models.

All the models were implemented in \textsc{Python} 3.10.10. All the CP models were solved with IBM ILOG CP Optimizer \cite{Laborie2018} from \textsc{CPLEX} 22.1.1, using its \textsc{Python} interface \cite{ibm2023cpoptimizerpython}. A time limit of 10 minutes (i.e., 600 seconds) was imposed on all the CP models in the small-to-medium instances of  \S~\ref{sec: initial comparison}, because a scheduling system in several areas of a wafer fab is expected to generate a Gantt-chart schedule every few minutes \cite{Ham2017-CP-p-batch-incompatible}. On the other hand, a time limit of 20 minutes (i.e., 1,200 seconds) was enforced on the large-scale instances of \S~\ref{sec: large-scale results} due to the increased scale. The RP model was solved with \textsc{Gurobi Optimizer} version 12.0.0 \cite{gurobi} with a time limit of one hour in an attempt to find optimal solutions. All the experiments were run on the PACE Phoenix cluster \cite{PACE}, using machines that run Red Hat Enterprise Linux Server release 7.9 (Maipo) with dual Intel\textregistered~ Xeon\textregistered~ Gold 6226 CPU @ 2.70GHz processors, with 24 cores, and 48 GB RAM, and parallelizing up to three experiments at the same time. Each run uses 8 cores and 16 GB RAM.

\subsection{Small-to-medium instances} \label{sec: initial comparison}

This section presents an initial comparison of the proposed s-A and s-A* models against the existing models in the literature: the RP MIP model by \citet{Shahvari2017-bLB} (formulation at \cite[Appendix~A]{Huertas2025_sbatching_ORP}) and the IA, G, and H CP models by \citet{Huertas2025_sbatching_ORP}.

The comparison is performed on the exact same $1{,}170$ small-to-medium instances introduced in \citet{Huertas2025_sbatching_ORP}, which were designed to gradually increase in size. Across these instances, the number of jobs, families, and machines ranges from $15$ to $100$, $2$ to $7$, and $2$ to $5$, respectively. Processing times and job weights are drawn from discrete uniform distributions, while family-dependent setup times are first generated as random values scaled in $[0,S]$ with $S \in \{20,50,100\}$, corresponding to loosely, moderately, and tightly loaded settings as in \citet{Shahvari2017-bLB}, and then adjusted to satisfy the triangular inequality. Release times are generated at random based on a lower bound on the makespan, so that jobs may arrive throughout the planning horizon. Full details of the instance generation procedure are provided in \citet{Huertas2025_sbatching_ORP}.

Minimum batch sizes $l_f$ are defined exactly as in \citet{Huertas2025_sbatching_ORP}. Each instance is first solved with the Core Model~\ref{model:core} to obtain a baseline schedule, from which the minimum number of consecutive jobs of each family $f \in \Families$ is extracted. The corresponding minimum batch size $l_f$ is then chosen at random between this value plus one and $|\Jobs_f|$. This construction ensures that the Core model’s schedule becomes infeasible under the minimum batch-size requirements and thus motivates the use of extended CP models, such as IA, G, H, s-A, and s-A*, that explicitly enforce these constraints.

\subsubsection{Which model is better?}

\begin{figure*}[t!]
    \centering
    \includegraphics[width=\linewidth]{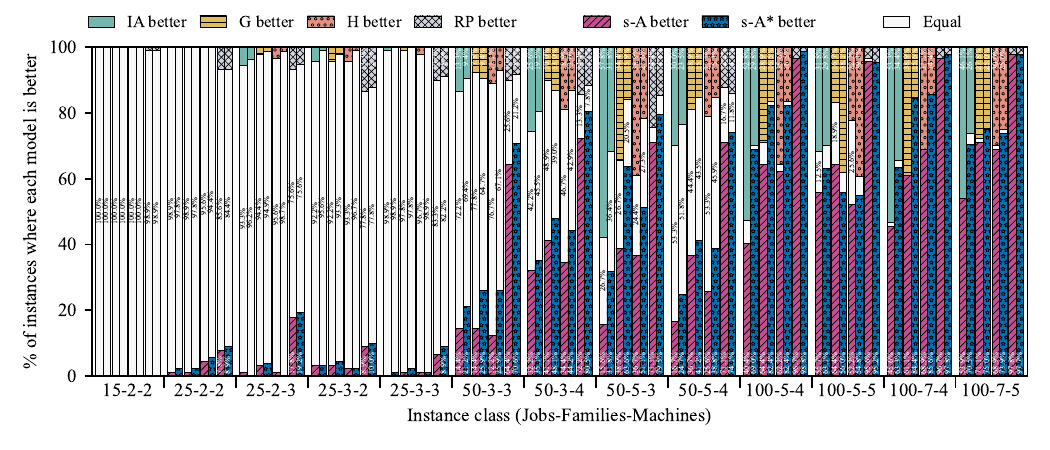}
    \caption{Percentage of small-to-medium instances where each proposed model is better than existing models}
    \label{fig:better models small}
\end{figure*}

Figure~\ref{fig:better models small} compares the proposed s-A and s-A* models against all existing CP models and the RP model. The horizontal axis groups the $1{,}170$ instances into 13 classes of 90 instances each, according to their number of jobs, families, and machines (J–F–M). For each class, there are eight columns. Each pair of adjacent columns compares the \emph{new} models (s-A and s-A*) against the same \emph{old} model (IA, G, H, or RP), which makes it possible to visually compare not only the new models against the old ones, but also allows the comparison between the new models under the same conditions. Within each column, three stacked bars are used: the first (bottom) bar shows the percentage of instances in which the new model yields a better objective value (lower TWCT) than the corresponding old model; the second (middle) bar shows the percentage of instances where both models obtain the same objective value; and the third (top) bar shows the percentage of instances in which the old model is better than the new one.

The trend is clear. For instances with up to 25 jobs, the proposed models perform essentially on par with the existing ones, as indicated by the dominance of the middle (white) bars. As the instance size grows, however, the bottom bars associated with the proposed s-A and s-A* models consistently dominate, showing that they outperform the previous models on an increasing fraction of instances. It is also evident that, when compared against the same existing model, the blue starred bar, associated with the s-A* model, is always larger than the magenta-striped bar, associated with the s-A model. This indicates that the additional formulation refinements and search phases in the s-A* model systematically strengthen the baseline s-A model.

\subsubsection{How much better?}

\begin{figure}[t!]
    \centering
    \includegraphics[width=\linewidth]{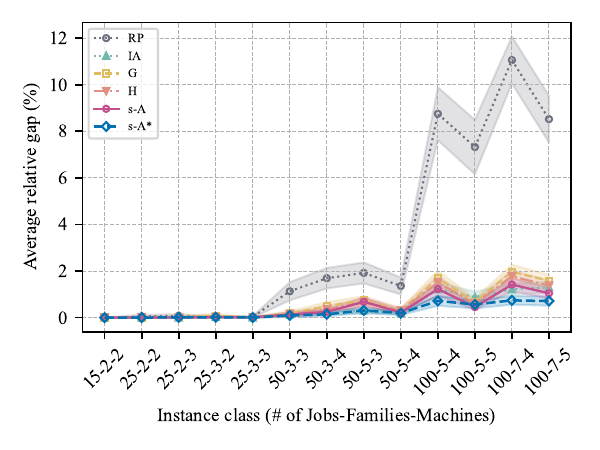}
    \caption{Average relative gaps}
    \label{fig:relative gaps small}
\end{figure}

Figure~\ref{fig:better models small} shows that the proposed CP models outperform the existing CP and MIP models on a larger fraction of instances. Figure~\ref{fig:relative gaps small} complements this by quantifying \emph{how much} better they are using the relative gap of each model. For each instance $i$ and model $\mathfrak{m}$, the relative gap is computed as
\[
    gap_i(\mathfrak{m}) = \frac{\lvert TWCT_{\mathfrak{m}}(i) - TWCT^*(i) \rvert}{\lvert TWCT_{\mathfrak{m}}(i) \rvert},
\]
where $TWCT_{\mathfrak{m}}(i)$ is the TWCT obtained by a given model $\mathfrak{m}$ on instance $i$ and $TWCT^*(i)$ is the best TWCT obtained for instance $i$ across all models that solved it. The figure reports, for each instance class, the average relative gap of each model together with its 95\% confidence interval.

The trend is again clear. For instances with up to 25 jobs, all models exhibit essentially identical average gaps. From 50 jobs onward, the RP model shows the largest average gaps across all classes. The IA, G, and H models remain very close to each other, while the s-A model already achieves slightly smaller gaps. Overall, the s-A* model consistently attains the smallest average relative gaps among all models.

\subsubsection{Comparison against MIP}

\begin{figure}[t!]
    \centering
    \includegraphics[width=\linewidth]{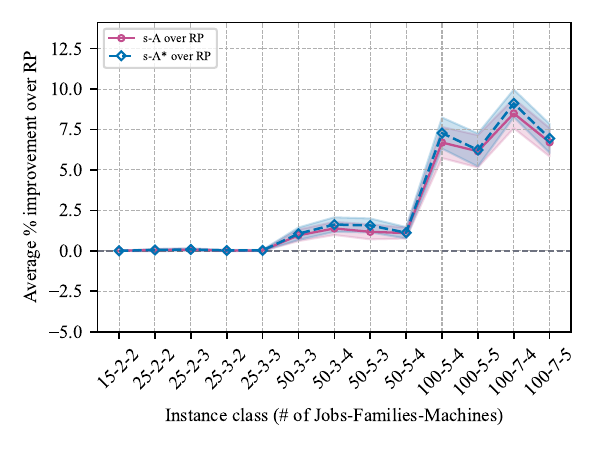}
    \caption{Average percentage of improvement of the s-A and s-A* models over RP}
    \label{fig:avg improvement small}
\end{figure}

Although the average relative gaps already show that the s-A and s-A* models dominate the existing ones, this subsection focuses more specifically on their improvement over the RP model. To do so, this subsection reports on the average percentage improvement (API) of the proposed CP models over the MIP model. The percentage improvement (PI) of a model $\mathfrak{m}_1$ over a model $\mathfrak{m}_2$ on a single instance $i$ is defined as
\begin{equation*}
    PI^{\mathfrak{m}_1}_{\mathfrak{m}_2}(i) = 100 \times \frac{TWCT_{\mathfrak{m}_2}(i) - TWCT_{\mathfrak{m}_1}(i)}{TWCT_{\mathfrak{m}_2}(i)},
\end{equation*}
For each instance class, the API is then computed as the average PI over its 90 instances. Figure~\ref{fig:avg improvement small} reports the API of the proposed models over the RP model across all 13 instance classes. For instances with up to 25 jobs, the proposed CP models obtain the same objective values as the MIP model. For larger instances, the proposed models consistently outperform the RP model, reaching improvements of more than 7\% on classes with up to 100 jobs. A key observation is that the RP model is given a time limit of one hour, whereas the CP models achieve these improvements within only 10 minutes.

Although this paper does not explicitly compare the proposed models against the existing TS meta-heuristic in the literature for s-batching with minimum batch sizes \cite{Shahvari2017-bLB}, they enable an indirect qualitative comparison, which was not addressed by \citet{Huertas2025_sbatching_ORP}. \citet{Shahvari2017-bLB} evaluated their TS on only 22 instances, with the largest instance being a moderately loaded instance containing 27 jobs. On that instance, the TS required 3 hours to improve the RP solution by about 7\%. In contrast, the present paper evaluates s-A and s-A* on more than one thousand instances between loosely, moderately, and tightly loaded instances with up to 100 jobs, and achieves improvements of a similar magnitude over the RP model, but within just a 10-minute time limit.

\subsection{Large-scale instances} \label{sec: large-scale results}

The previous section focused on small-to-medium instances with up to 100 jobs, 7 families, and 5 machines. This new section uses 3,750 new large-scale instances where the number of jobs can take five possible values up to 500, i.e., $|\Jobs| \in \set{100, 200, 300, 400, 500}$, and the number of families and machines can also take five possible values up to 10, i.e., $|\Families|, |\Machines| \in \set{4,5,6,8,10}$. The setup times, processing times, job weights, release times, and minimum batch sizes were computed in the same way as in \S \ref{sec: initial comparison}. For each combination of the possible values for the number of jobs, families, machines, and setup times distributions, a total of 10 instances were generated, resulting in $5 \times 5 \times 5 \times 3 \times 8 = 3,750$ instances in total. To put the scale of these instances in perspective, the simulation models in the SMT 2020 start no more than 10k wafers per week \cite{SMT2020-Paper}, which would correspond to 400 jobs, assuming that each job is a wafer lot with 25 wafers. Thus, the new large-scale instances used in this paper are comparable to the only academic instances known by the authors that are close to real-world scenarios.

This section also explores the impact of the inference level of the constraint propagators, which determines how aggressively the solver applies constraint propagation to reduce the search space. IBM ILOG CP Optimizer \cite{Laborie2018} supports several inference levels: \textit{low}, \textit{basic} (default), \textit{medium}, and \textit{extended}. The \textit{low} level applies minimal propagation to reduce runtime per node, whereas the \textit{extended} level applies more sophisticated reasoning and consistency enforcement, which can be beneficial in highly constrained models but may also increase propagation overhead and overall solution time due to more expensive consistency checks.

Since the previous section already demonstrated that the RP model is not competitive when compared to the existing CP approaches, it was not considered on the large-scale instances. Instead, each of the $3{,}750$ large-scale instances was solved with seven CP models: IA, G, H, s-A, s-A(M), s-A*, and s-A*(M), where the suffix (M) indicates that CP Optimizer is run with the \textit{medium} inference level for the constraint propagators instead of the default. The following subsections report on the models' capacity to find solutions, and then it dives into the proposed models' capacity to find better solutions than the existing approaches, closing with the average improvement of the proposed models over the existing ones. To be able to assess the impact of the instance size as it grows, Figures~\ref{fig: percentage solved}-\ref{fig: average percentage improvement large} show stacked line or bar charts, each displaying the results of the instances with s specific number of jobs, visible on the right vertical axis. The horizontal axis groups instances depending on their number of families and machines.

\subsubsection{Capacity to find solutions}

\begin{figure}[t!]
    \centering
    \includegraphics[width=\linewidth]{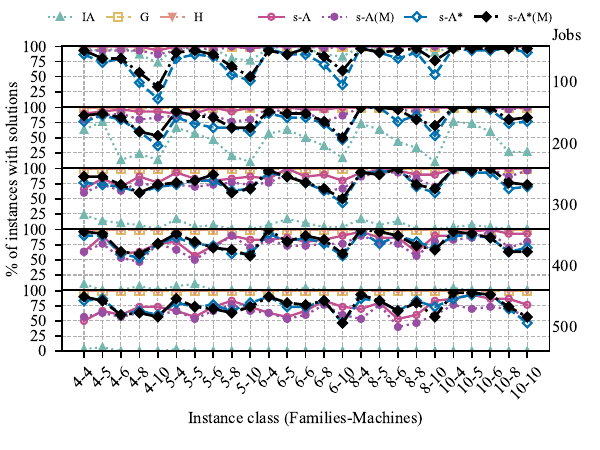}
    \caption{Percentage of instances where each model found a solution}
    \label{fig: percentage solved}
\end{figure}

Figure~\ref{fig: percentage solved} reports, for each model, the percentage of instances on which it was able to find a solution. Each combination of Jobs (vertical right axis) and families-machines (horizontal axis) aggregates the results of 30 runs (instances), and each line in the graph corresponds to one of the seven models compared.

\citet{Huertas2025_sbatching_ORP} reported that, on the small-to-medium instances, the IA model performed best among the CP approaches on the IPF variation, mainly due to the sum-of-presences constraints used to enforce minimum batch sizes. However, these constraints operate over binary presence variables, which become a bottleneck for the search on large-scale instances. Figure~\ref{fig: percentage solved} clearly shows that, as instance size grows, the IA model quickly loses robustness and eventually fails to even find solutions. By contrast, the G and H models remain able to find solutions across all large-scale instances, which can be attributed to their additional global constraints that strengthen propagation and help the search.

Figure~\ref{fig: percentage solved} also shows that the proposed s-A variants (i.e., the s-A, s-A(M), s-A*, and s-A*(M) models) do not solve \emph{all} the large-scale instances. In fact, they solved on average 81.4\% of them. Naturally, the most challenging instances are those with the most number of machines, where the combinatorial decisions grow exponentially. When examining the impact of the inference levels, Figure~\ref{fig: percentage solved}, shows that the s-A*(M) series is usually at par or slightly above the s-A* series. However, this pattern is not present between the s-A and s-A(M) models, where the baseline model (i.e., s-A) is usually above its counterpart with strengthen inference levels (i.e., s-A(M)). This suggest that the strengthen inference levels are useful to find solutions only in combination with targeted search phases.

\subsubsection{Finding better solutions than existing methods}

\begin{figure*}[t!]
    \centering
    \includegraphics[width=\linewidth]{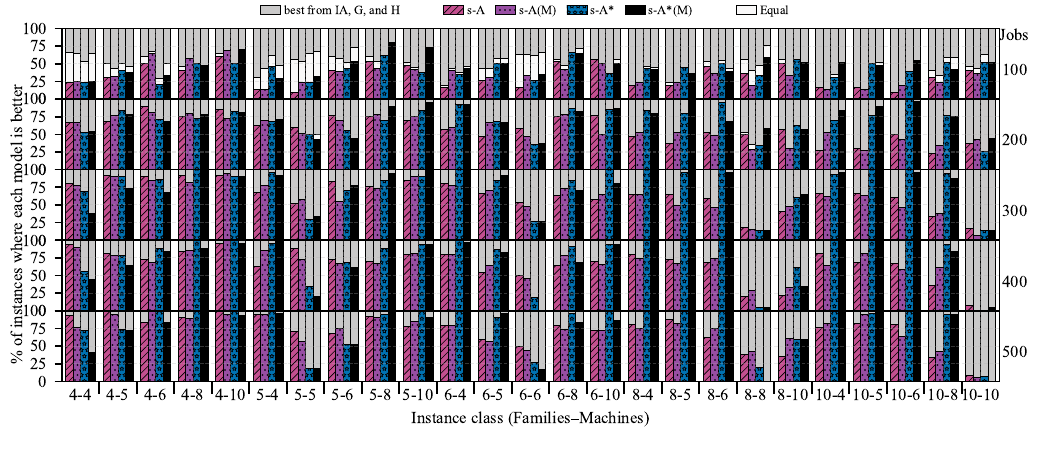}
    \caption{Percentage of large-scale instances where each proposed model is better than existing models}
    \label{fig:better models large}
\end{figure*}

\textit{Although Figure~\ref{fig: percentage solved} shows that the proposed CP models do not solve every large-scale instance, they outperform the existing methods on the instances for which they do find solutions.} Figure~\ref{fig:better models large} shows, for these instances, the percentage of cases where each one of the proposed models outperforms the \emph{best} among the IA, G, and H models. The figure follows the same structure as Figure~\ref{fig:better models small}, but instead of comparing every proposed model against every existing method, for each individual instance, the best result obtained across the IA, G, and H models is saved as the best solution from an existing model. For each job-size group (shown on the right vertical axis) and each families–machines class (on the horizontal axis), four columns compare the four proposed models against the result stored as the best from the existing models. Within each column, three stacked bars are used: the bottom bar shows the percentage of instances where the proposed model achieves a better objective value (lower TWCT), the top bar shows the percentage of instances where the result stored as best of the existing models is better than the proposed model, and the middle (white) bar shows the percentage of instances where both models obtain the same objective value. To ensure a fair comparison, all the percentages in a given column are computed only over instances where the two models compared in such column found a solution.

The trends are consistent with the small-to-medium case. As the number of jobs increases, the proposed models outperform the existing CP models on a growing fraction of instances. However, another trend reveals that instances where the number of jobs and machines coincide are the most difficult ones for the proposed models. This can be explained by the increased symmetries on these instances, and the lack of symmetry-breaking constraints that prune them. Finally, the effect of the medium inference level is barely noticeable The medium inference level is barely noticeable and actually detrimental in most cases. In the instances with 100 jobs, the strengthen inference level is actually useful only on instances with up to 5 families. However, as the number of families grows, its effect fades away and even becomes detrimental. This also happens when the number of jobs increases, making the default s-A* model better in most of the cases, with counted exceptions due to the randomness of the instances. This can be explained because the strengthen inference levels require additional time to propagate constraints and find better solutions. Instead, the default inference is faster in nature,which allows faster exploration.

\subsubsection{Improvement over existing models}

\begin{figure}[t!]
    \centering
    \includegraphics[width=\linewidth]{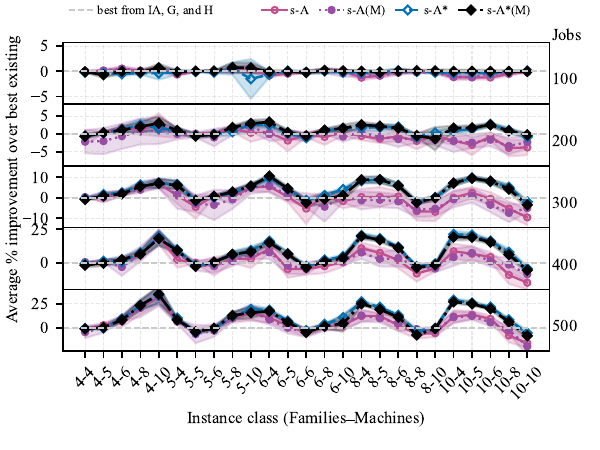}
    \caption{Average percentage of improvement (API) of the proposed models over the existing ones}
    \label{fig: average percentage improvement large}
\end{figure}

While Figure~\ref{fig:better models large} focuses on how often the proposed models outperform the existing CP models, Figure~\ref{fig: average percentage improvement large} quantifies \emph{by how much} they improve the objective value. The figure reports, for each instance class, the (API) of each proposed model over the best among the IA, G, and H models, with their 95\% confidence intervals. This comparison can only be made on individual instances where the both (i) the proposed model compared and (ii) the best of the existing models found a solution.

The trends are clear. For instances with 100 jobs, the APIs of all proposed models are very close to zero, confirming that on this scale the Aligned models match the best existing CP models in solution quality. \textit{The key insight from Figure~\ref{fig: average percentage improvement large}, and perhaps of this paper, is that as the number of jobs increases, the APIs of the proposed Aligned models over the existing ones in the literature steadily grow, reaching improvements of up to 25\% on classes with 500 jobs}. 

Nonetheless, the trends in Figure~\ref{fig: average percentage improvement large} further confirm that the most challenging instances for the proposed models are those where the number of families and machines are the same, i.e., $|\Families| = |\Machines|$, which is the case where more symmetries appear. Early attempts to include symmetry-breaking constraints included fixing a 1:1 relationship between families and machines on such instances, but this approach was ruled out as it prunes improving solutions that exist due to the non-identical release times. Finding good-quality symmetry breaking constraints with such a tight and compact formulation is still a potential research stream to further improve the aligned models.

Figure~\ref{fig: average percentage improvement large} also demonstrate the benefit of the improved formulation with tailored search phases. The s-A and s-A(M) models, which do not have the enhanced formulation nor tailored search phases, consistently show a detrimental API below 0\% on instances with 200 and 300 jobs and more than 6 families. However, on instances with 400 and 500 families, their API steadily increase above 0\%, except on instances with more than 8 machines. In contrast, the s-A* and s-A*(M) models, which benefit from an improved formulation and customized search phases, consistently present a greater API, which is always above 0\%, except on instances with an equal number of families and machines. \textit{These trends suggest that beyond 500 jobs, the proposed CP models are expected to perform even better than the existing CP models in the literature; and to allow them find solutions on \textit{all} instances, it would be necessary to increase the time limit beyond 20 minutes.} Finally, even though the strengthen inference level of the constraint propagators help the s-A*(M) find solutions to more large-scale instances, its impact on the solution quality is barely noticeable, with the series of the s-A and s-A(M) models moving close together, and the same happening with the series of the s-A* and s-A*(M) models.

\section{Concluding remarks} \label{sec: conclusion}

This paper proposed an \textit{Aligned} CP model for serial batch scheduling with minimum batch sizes, namely the s-A model. Unlike existing CP models in the literature, the s-A model does not rely on a virtual set of possible batches, which adds complexity to the problem structure and exponentially increases the number of variables. Instead, it reasons directly on consecutive jobs of the same family by constructing \emph{family blocks}. To do so, additional virtual interval variables are introduced for each job that span the entire duration of its family block. In this way, the virtual intervals of all the jobs in the same family block are aligned, effectively creating a parallel batch. A cumulative function defined over these overlapping virtual intervals counts how many jobs belong to each family block, and bounds on this function enforce the minimum and maximum batch-size requirements. An improved formulation that exploits the structure of the problem and adds targeted search phases results in the s-A* model, which further strengthens propagation on the core job intervals.

Extensive experiments on over one thousand small-to-medium instances with up to 100 jobs show that the proposed Aligned models consistently outperform existing methods, including the RP MIP model, the TS metaheuristic of \citet{Shahvari2017-bLB}, and the IA, G, and H CP models of \citet{Huertas2025_sbatching_ORP}. On these instances, the Aligned models match the performance of existing approaches on the smallest cases and clearly dominate as instance size grows, achieving improvements of more than 7\% over the RP model within a 10-minute time limit. Additional experiments on over three thousand large-scale instances with up to 500 jobs, 10 families, and 10 machines analyze the impact of stronger inference levels in the propagators. Although the proposed models do not solve every large-scale instance within a time limit of 20 minutes, whenever they find solutions (on around 76, those solutions are significantly better than those obtained by the existing CP models, with average percentage improvements reaching up to 25\% over the best among the IA, G, and H models.

These results demonstrate that high-quality schedules can be obtained in under 20 minutes for instances of realistic wafer-fab scale, with up to 500 jobs. This level of performance makes CP-based serial batching a promising option for fast rescheduling in semiconductor manufacturing environments.

Ongoing work includes designing new CP models for s-batching with minimum batch sizes that are compatible with open-source solvers. Future work includes embedding both serial and parallel Aligned models into richer wafer-fab scheduling and digital-twin frameworks that couple CP-based scheduling with detailed discrete-event simulation.

\bibliographystyle{IEEEtranN}

\bibliography{references}

\end{document}